\documentclass{isprs} 
\usepackage{subfigure}
\usepackage{setspace}
\usepackage{geometry} 
\usepackage{epstopdf}
\usepackage[labelsep=period]{caption}  
\usepackage[british]{babel} 
\usepackage[hang]{footmisc}

\usepackage[authoryear]{natbib}

\usepackage{amssymb}
\usepackage{booktabs}
\usepackage{lmodern}
\usepackage{multirow}
\usepackage{utfsym}
\usepackage{url}
\usepackage{xcolor}

\usepackage{xspace}
\makeatletter
\DeclareRobustCommand\onedot{\futurelet\@let@token\@onedot}
\def\@onedot{\ifx\@let@token.\else.\null\fi\xspace}

\def\ie{\emph{i.e}\onedot}

\makeatother

\begin{document}

\title{A Low-Cost Portable Lidar-based Mobile Mapping System\\ on an Android Smartphone}
\date{Feb 10 2025}


\author{
Jianzhu Huai\textsuperscript{1}, Yuxin Shao\textsuperscript{1}, Yujia Zhang\textsuperscript{2}\thanks{Corresponding author} , %
Alper Yilmaz\textsuperscript{3}}

\address{
\textsuperscript{1 }State Key Lab of Info Engineering in Surveying, Mapping, and Remote Sensing, Wuhan University, \\129 Luoyu Road, Wuhan, Hubei, China \\
\textsuperscript{2 }School\ of Instrument Science and Engineering, Southeast University, \\Sipailou, Nanjing, 210096, Jiangsu, China - yujia\_zhang@seu.edu.cn\\
\textsuperscript{3 }Dept.\ of Civil, Environmental and Geodetic Engineering, The Ohio State University, \\Columbus, OH 43210, USA 
\\
}



\abstract{
The rapid advancement of the metaverse, digital twins, and robotics underscores the demand of low-cost, portable mapping systems for reality capture. Current mobile solutions, such as the Leica BLK2Go and lidar-equipped smartphones, either come at a high cost or are limited in range and accuracy. 
Leveraging the proliferation and technological evolution of mobile devices alongside recent advancements in lidar technology, we introduce a novel, low-cost, portable mobile mapping system. Our system integrates a lidar unit, an Android smartphone, and a GNSS-RTK stick. Running on the Android platform, it features lidar-inertial odometry built with the NDK, and logs data from the lidar, wide-angle camera, IMU, and GNSS. With a total bill of materials (BOM) cost under \$2,000 and a weight of about 1 kilogram, the system achieves a good balance between affordability and portability. We detail the system’s design, multisensor calibration, synchronization, and evaluate its performance for 3D reconstruction, focusing on tracking and mapping accuracy. To further contribute to the community, the system's design and software has been made open source.
}


\maketitle


\section{Introduction}\label{sec:intro}
 
\sloppy

Service robots, immersive games, and digital twins all require rapid and accurate modeling of real-world scenes. Mobile mapping has grown in importance, especially in applications like automated construction, where it has proven highly valuable. With decreasing costs and shrinking form factors of lidars, many portable mapping systems have emerged, such as Hovermap and Leica BLK2GO. However, these systems often face limitations in terms of size, weight, and power (SWaP), and costs.

One major reason for the high costs of these devices is their reliance on customized displays and computing units. In contrast, Android phones serve as an ideal example of mobile computing, dominating a large share of the mobile device market. Driven by strong consumer demand, Android phones continually advance, introducing features such as wireless debugging, Ethernet tethering, high-resolution touchscreens, built-in batteries, wireless connectivity, and wide-angle cameras ($\ge$90$^\circ$). Additionally, rapid technological advancements have led to significant e-waste, as many used phones are discarded. We believe that developing a mobile mapping system on an Android phone can not only reduce the bill of materials (BOM) but also offer an environmentally friendly solution.

\begin{figure}[!htbp]
\includegraphics[width=0.45\columnwidth]{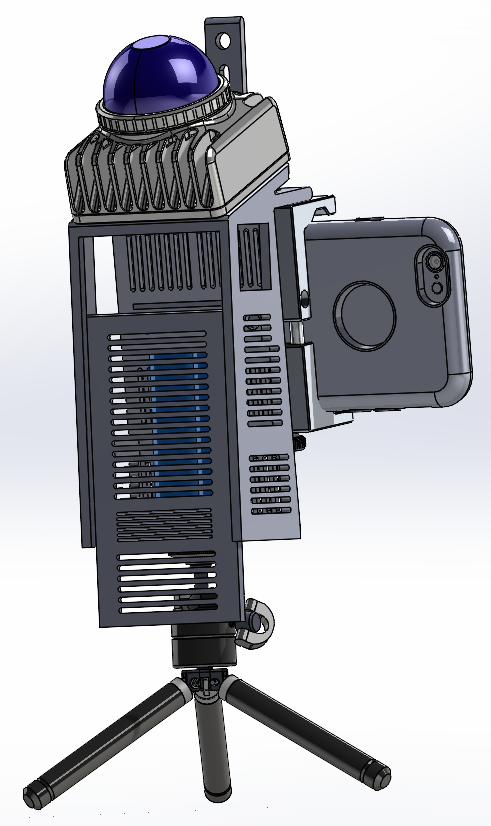}
\includegraphics[width=0.4\columnwidth]{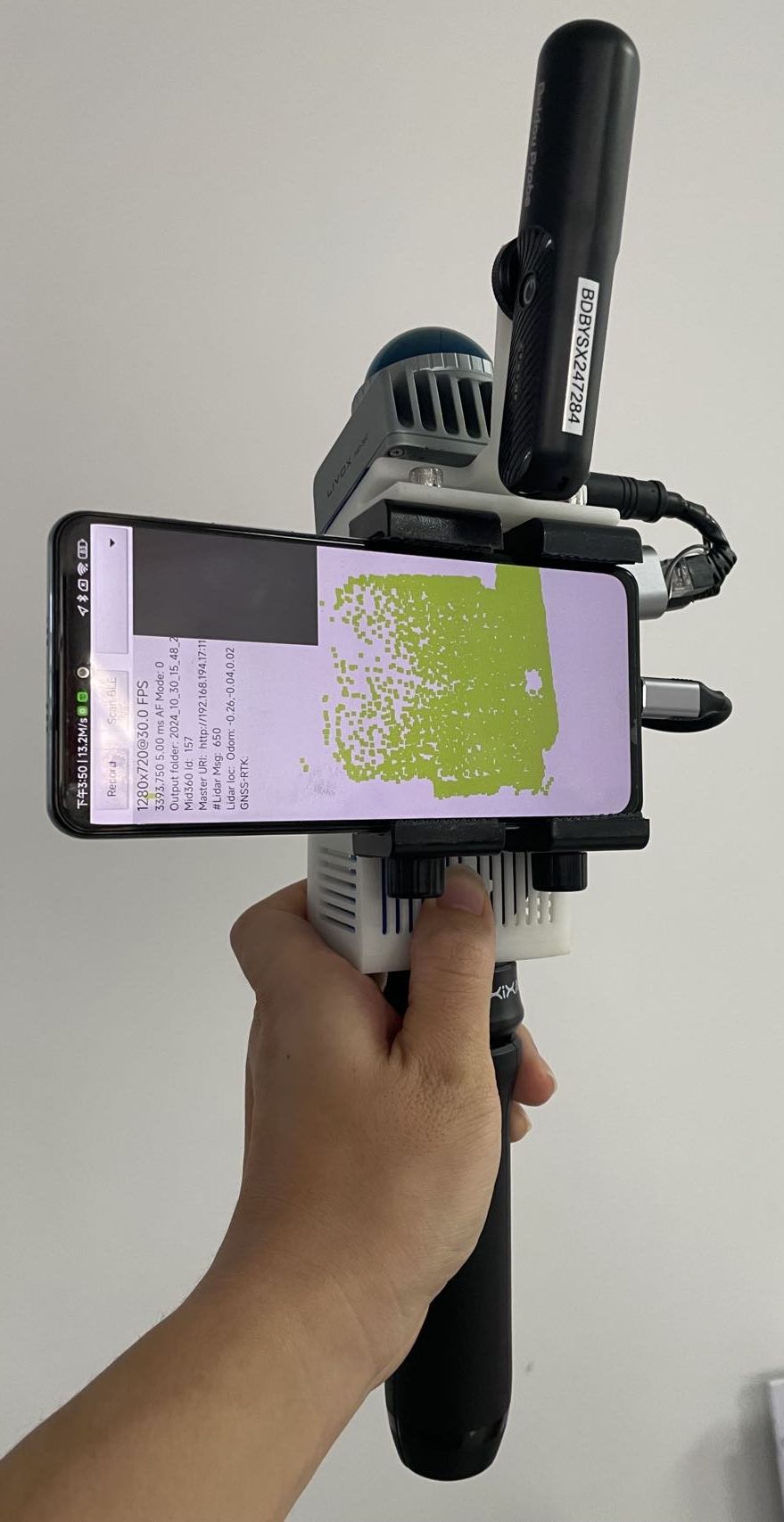}
\caption{
The CAD model (left) and prototype (right) of the Livox-Phone mobile mapping system.}
\label{fig:prototype}
\end{figure}

With this goal in mind, we develop a mobile mapping system based on the Android platform.
Our prototype (Fig.~\ref{fig:prototype}) integrates a low-cost lidar with a medium-cost Android phone and includes a GNSS-RTK module for accurate georeferencing. 
Leveraging the Android phone’s built-in capabilities—computing power, touchscreen, camera, and wireless communications (Bluetooth and 4G),
the system offers a compact design with a cost below \$2000, weighing approximately 1 kg, making it suitable for handheld operation of extended periods.
To our knowledge, this is the first kind of mobile mapping system based on a smartphone with a long range lidar.

On the software side, our system features a logging module that records data from the camera, IMU, GNSS-RTK, and lidar, as well as a lidar-inertial odometry (LIO) module for real-time pose tracking and mapping. The system is developed in Java using the Android API and Java native interface (JNI). For performance, message passing and LIO are implemented in C++ using ROS1 packages, compiled with the native development kit (NDK).

To show the usefulness of the system,
we design a calibration procedure to synchronize and calibrate the multisensor system.
By experiments, we evaluate the calibration quality, computation efficiency, and the tracking and mapping accuracy with the captured data online and offline.

The following sections are organized as follows. Section~\ref{sec:related} reviews related portable mobile mapping systems. Section~\ref{sec:method}
details the system design, implementation, and calibration procedures. In Section~\ref{sec:experiments}, we present the calibration evaluation and quantitative results for tracking and mapping. Finally, Section~\ref{sec:conclusions} summarizes our findings.

\section{Related Work}\label{sec:related}
In this section, we briefly overview existing mobile mapping systems utilized for 3D reconstruction and mapping purposes. These systems leverage a variety of sensor technologies, including laser scanning, visual odometry and integrated GNSS/IMU solutions. Table \ref{tab:related_work} present comparative analysis of representative mobile mapping systems alongside our proposed system.

\begin{table*}[!htbp]
\centering
\begin{tabular}{l|c|c|c|c|c|c|c}\toprule
Product &Year &Lidar type &Camera type &Cost &Weight &Display &GNSS-RTK \\ \midrule
Leica BLK2GO &2019 &Laser scanner &3 cameras &\$55635 &775 g &Desktop & \usym{2613} \\ \midrule
Wildcat &2022 &Velodyne VLP16 &4 cameras&\textgreater \$4000 &- &Desktop & \usym{2613} \\ \midrule
iPad mapper &2023 &Flash lidar sensor &1 camera &\$1299 &685 g &iPad & \usym{2613} \\ \midrule
OR-LIM &2024 &Velodyne VLP16 &1 fisheye &\textgreater \$4000 &- &Desktop & \usym{2613} \\ \midrule
3D Helmet &2024 &Mid360 &2 OAK-D Pro W &\textgreater \$2000 &- &Desktop & \usym{2613} \\ \midrule
Chcnav RS10 &2024 &Pandar XT16 alike&3 cameras &\$14000 &1.9 kg &Desktop & \checkmark \\ \midrule
Share SLAM S10 &2024 &Mid360 &2 wide-angle &\$8000 &1.01 kg & Phone & \checkmark\\ \midrule
Ours &2024 &Mid360&1 wide-angle&\$2000 &1 kg & Phone& \checkmark \\ 
\bottomrule
		\end{tabular}
	\caption{A summary of some related mobile mapping systems.}
\label{tab:related_work}
\end{table*}

A variety of platforms have been researched over the years.
Wildcat \citep{ramezaniWildcatOnlineContinuousTime2022} introduced a 3D lidar-inertial SLAM system featuring a rotating 3D lidar on a motor. 
The system later evolves into the commercial product Hovermap STX \footnote{\url{https://emesent.com/emesent-product/hovermap-series/}} for accurate mapping.
OR-LIM \citep{congLIMObservabilityawareRobust2024} presents a rotating lidar-based mapping system, incorporating a lidar-inertial odometry (LIO) module with an efficient surfel-map smoothing (SMS) module.
In contrast, \cite{cuiALiDAR2024} introduced a 3D mapping system in which both the LiDAR and IMU are mounted on a motor, rotating together as a unit.
Meanwhile, \cite{hamesseDevelopmentUltraportable3D2024} 
proposed a helmet-mounted mobile mapping system with a Livox Mid360 lidar.
Additionally, the iPad mapper \citep{teoEvaluatingAccuracyQuality2023} utilizes the iPad Pro’s lidar sensor to collect data under various scanning conditions, and assesses the sensor’s accuracy and potential for generating 3D BIM models.

Other studies, such as \cite{degeyterPointCloudValidation2022}, have compared various data acquisition techniques within laser scanning processes, with a particular emphasis on their semantic segmentation capabilities for BIM modeling. 
In \cite{huRobotassistedMobileScanning2023}, a novel robot-assisted mobile laser scanning approach is introduced, integrating SLAM algorithms with legged robot motion control and pathfinding for automated 3D reconstruction and point cloud semantic segmentation. \cite{luoIndoorMappingUsing2023} proposes an automated and efficient indoor mapping method that utilizes low-cost mobile laser scanner (MLS) point clouds, enhanced by architectural skeleton constraints. Meanwhile, \cite{pantoja-roseroImagebasedGeometricDigital2023} employs five synchronized cameras in an image-based pipeline for creating geometrical digital twins (GDTs) of stone masonry elements. Lastly, \cite{wangDevelopmentLowcostVisionbased2023} introduces a fast image processing software designed for single-board computers (SBCs), utilizing a low-cost laser rangefinder, camera, and Raspberry Pi for civil structural health monitoring.

\begin{figure}[!htbp]
\includegraphics[width=\columnwidth]{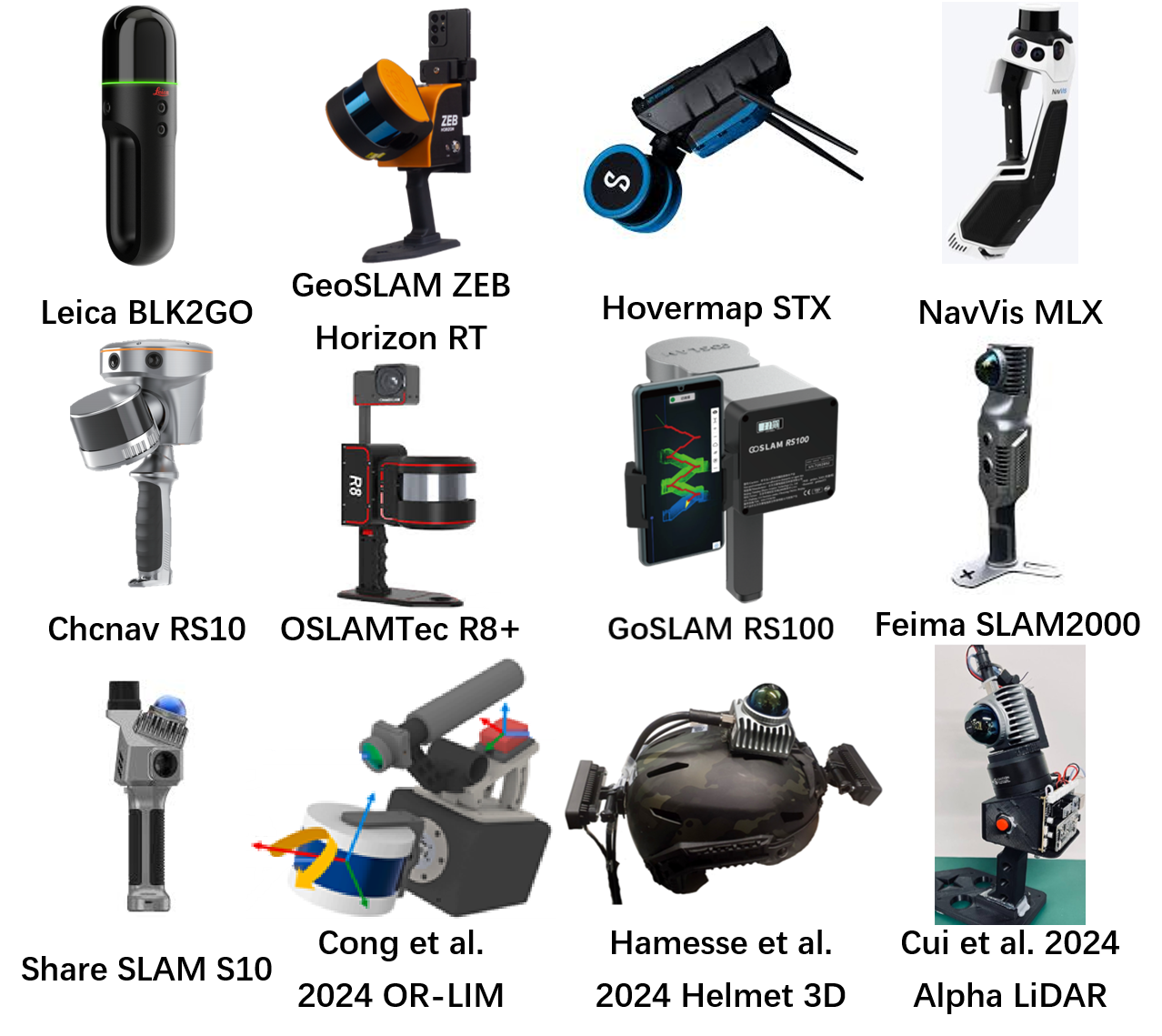}
\caption{Hardware designs of related portable mobile mapping systems.
}
\label{fig:mms}
\end{figure}

In addition to systems developed for research, there are many consumer-grade, assembled mobile mapping systems available (Fig.~\ref{fig:mms}). For instance, the Leica BLK2GO\footnote{\url{https://shop.leica-geosystems.com/leica-blk/blk2go}} is a handheld imaging laser scanner that creates a 3D digital twin as users walk through an area. The CHCNAV RS10\footnote{\url{https://www.chcnav.com/product-detail/rs10}} integrates GNSS-RTK, laser scanning, and visual SLAM into a single platform that enhances the efficiency and accuracy of indoor and outdoor 3D scanning and surveying. The Share SLAM S10\footnote{\url{https://www.shareuavtec.com/S10}} features a 20-degree tilted lidar setup, dual cameras with a 135$^\circ$ field of view (FOV), and an RTK positioning module. Also, this system enables real-time previewing of point clouds on smartphones.

To our knowledge, these products are often too costly to average mapping users, limiting their availability. Moreover, unlike the above systems, our design leverages the smartphone for data processing and visualization, thereby reducing the material cost.

\section{Livox Phone for Mobile Mapping}\label{sec:method}
This section describes our system design, implementation, and the calibration and synchronization procedures.

\subsection{Hardware}
The proposed mobile mapping system comprises a casing, an Android phone, a Livox Mid360 lidar, a GNSS-RTK stick, and a tripod. The casing houses the lidar battery, connectors, and a switch. The Android phone is chosen to be a mid-range model for a good balance between computing power and cost.
The GNSS-RTK stick, sold by Woncan in Wuhan, China, connects to the phone via Bluetooth. The lidar is linked to the phone through an RJ45-to-Type-C converter. All components are securely attached with standard screws as needed.

The Mid360 lidar was selected for its RJ45 connector, ability to modify its static IP address to a specific subnet, and open-source driver, making it compatible with Android via NDK. An alternative lidar model with similar features, like the Ouster OS2, can also be integrated, since it has the RJ45 connector and an open-source driver (so we can port it to Android easily).
The Mid360 includes an internal IMU, which we use in subsequent tests. Alternatively, the smartphone’s internal IMU can serve as a backup when the lidar does not have a built-in IMU.

In our design, the Mid360 lidar of 360$^\circ\times$59$^\circ$ FOV is angled  downward to increase its overlap with the smartphone’s rear camera. 
The smartphone is also slightly tilted to capture more of the floor area. For reference, the Redmi K60 Pro's rear camera has a FOV of 72$^\circ\times$45$^\circ$, while the Samsung Galaxy S22+ offers a wider FOV of 102$^\circ\times$70$^\circ$.

\subsection{Software}
The system’s software consists of several core modules that support an Android app. These modules include lidar-inertial odometry, a multisensor logger, ROSJava for message passing, and screen display.
For real-time motion tracking, we use Faster-LIO \citep{baiFasterLIOLightweightTightly2022} due to its efficiency (see Section~\ref{sec:experiments}). ROSJava facilitates message passing, ensuring easy message handling and extensibility.

The multisensor logging module captures data from all sensors, including the GNSS-RTK, camera, lidar, and IMUs. Camera images at a resolution of 1280$\times$720 are recorded at 30 Hz, while the phone’s IMU data are logged at 200 Hz. The Mid360 IMU data are also recorded at 200 Hz, with lidar data logged at 10 Hz.

All logged data are timestamped with both host times and sensor times.
Camera and phone IMU data are originally timestamped with system boot times \citep{huaiMobileARSensor2019}.
The intervals between consecutive camera and phone IMU system boot timestamps are remarkably regular, thus they can be considered sensor times.
Additionally, we record the Unix timestamps at the moment these data are received.
By retouching the Livox ROS driver2, we log both the unix time (host time) and sensor time for the Mid360 laser and IMU data.
With jittering host times and regular sensor times, we can use the convex hull algorithm \citep{zhangClockSynchronizationAlgorithms2002} to smooth out timestamp jitter in the local host time,  either in a causal manner or in batch processing.

The app was developed in Java on Android 11. C++ code for lidar data reading, recording, and processing is handled with ROS1 packages, built with NDK and accessed via ROSJava in Android. Image and point cloud display use the OpenGL-based renderers.


\subsection{Multisensor Calibration}\label{subsec:calib}
The calibration process consists of multiple steps: calibrating the smartphone’s wide-angle camera,
spatio-temporal calibration between the rolling shutter (RS) camera and the IMU, 
temporal calibration between the LiDAR and the IMU, 
and spatial calibration between the LiDAR and the camera.

Wide-angle camera calibration requires holding the device to capture images at a number of stationary poses (50 in our tests) in front of a calibration board, such as an AprilGrid. Static frames are then extracted from the recorded video by analyzing framewise optical flow. The Kalibr toolkit \citep{mayeOnlineSelfcalibrationRobotic2016} then processes these static frames to estimate the camera’s intrinsic parameters. By default, our app locks the camera focus at infinity, resulting in a fixed focal length.

To calibrate the RS camera and the IMU, we use the RS variant of Kalibr \citep{huaiContinuoustimeSpatiotemporalCalibration2022} to estimate both the extrinsic parameters and time offset between the two sensors. This approach uses continuous-time B-splines with 50 Hz pose knots to represent the trajectory, optimizing it along with the calibration parameters based on calibration board observations and IMU data. Several in-motion sequences, each lasting about 60 seconds, are recorded indoors by moving the device in front of the calibration board. Exposure time is locked at 5 ms to minimize motion blur.

For lidar-IMU calibration, we use the same data sequences as for the RS camera-IMU calibration and process the lidar data with KISS-ICP \citep{vizzo2023ral}. Then, we estimate the time offset and relative rotation using a correlation-based method. This method first estimates the time offset by correlating the norms of the IMU’s angular rates with those computed from the lidar odometry, then solves for the relative orientation using a truncated least squares approach.

For LiDAR-camera extrinsic calibration, we collect multiple stationary sequences with the device on a tripod positioned in environments rich in linear features. From each sequence, two seconds of LiDAR data are aggregated and paired with a corresponding image. These scan-image pairs are then processed using the livox\_camera\_calib tool \citep{yuanPixellevel2021} to estimate the LiDAR-camera extrinsic parameters.

\section{Experiments}\label{sec:experiments}
This section evaluates our system's calibration quality, tracking performance, and mapping accuracy. For our prototypes, we used two pre-owned smartphones: a Redmi K60 Pro (released in 2023) and a Samsung Galaxy S22+ (released in 2022).

\subsection{Calibration Evaluation}
We ran the calibration procedure described in Section~\ref{subsec:calib} and evaluated the calibration accuracy of parameters between sensors.

Five in-motion sequences were captured for each phone. To minimize sequence correlation, the app was terminated and restarted before each recording. 
To assess the repeatability of calibration, we collected two datasets for the S22+ device in January 2025 and February 2025.
We calculated the sample standard deviations of the estimated camera rotation relative to the Livox IMU, $\mathbf{R}_{IC}$; the camera position relative to the IMU, $\mathbf{p}_{IC}$; and the lidar rotation relative to the IMU, $\mathbf{R}_{IL}$. These results are listed in Table~\ref{tab:sigma_extrinsics}.

As shown in the table, the nonlinear optimization-based RS camera-IMU calibration produced very consistent results. Time offsets between the phone camera and Livox IMU showed good repeatability across different app runs. The correlation-based approach also yielded reliable rotation estimates. However, the time offsets between the Livox lidar and IMU exhibited slightly larger variations, though within 5 ms. Optimization-based calibration methods could further improve these results.
Despite our efforts to tune the lidar-IMU calibration tool \texttt{lidar\_IMU\_init} \citep{zhu2022robust}, it consistently aborted before completing extrinsic initialization on our sequences.

\begin{table}[!htbp]
\centering
\caption{
Sample standard deviations for the camera-IMU and lidar-IMU extrinsic calibrations. Here, $\mathbf{p}$ and $\mathbf{R}$ represent the relative position and rotation, respectively; $t_d$ denotes the time offset, and $d$ indicates the rolling shutter line delay.}
\label{tab:sigma_extrinsics}
\begin{tabular}{@{}c|cccc@{}}
\toprule
\begin{tabular}[c]{@{}c@{}}camera-\\ IMU\end{tabular} &
  \begin{tabular}[c]{@{}c@{}}$\sigma(\mathbf p_{IC})$\\ (mm)\end{tabular} &
  $\sigma(\mathbf R_{IC})(^\circ)$ &
  $\sigma(t_d) (\mu s)$ &
  \begin{tabular}[c]{@{}c@{}}$\sigma(d)$\\ $(\mu s)$\end{tabular} \\ \midrule
K60Pro    & 7.94 & 0.12                             & 675.53                & 2.15 \\ \midrule
S22+ Jan     & 1.16 & 0.05                             & 603.87                 & 0.27 \\ \midrule
S22+ Feb     & 0.51 & 0.02                             & 445.57                 & 0.38 \\ \midrule
lidar-IMU &      & $\sigma(\mathbf R_{IL})(^\circ)$ & $\sigma(t_d) (\mu s)$ &      \\ \midrule
K60Pro    &      & 0.44                             & 2624                  &      \\ \midrule
S22+ Jan      &      & 0.49                             & 3108                  &      \\ \midrule
S22+ Feb     &      & 0.85                             & 2265                  &      \\ \bottomrule
\end{tabular}
\end{table}

The mean time offsets between the camera and the Livox IMU, as well as between the lidar and the Livox IMU, computed from these motion sequences, are listed in Table~\ref{tab:time-offsets}. These offsets up to 20 milliseconds, show the necessity of temporal calibration for achieving accurate mapping.

\begin{table}[!htbp]
\centering
\caption{
Mean time offsets in milliseconds of the phone camera and Livox lidar relative to the Livox IMU for the K60Pro and S22Plus devices. The time offset is defined as the correction applied to the camera or lidar message timestamps.}
\label{tab:time-offsets}
\begin{tabular}{@{}cc|cc|cc@{}}
\toprule
\multicolumn{2}{c|}{K60Pro} & \multicolumn{2}{c|}{S22+ Jan} & \multicolumn{2}{c}{S22+ Feb} \\ \cmidrule(l){1-6} 
camera        & lidar       & camera      & lidar & camera      & lidar     \\ \midrule
9.59         & -9.04       & 20.40       & -8.81   & 19.60       & -11.13   \\ \bottomrule
\end{tabular}
\end{table}

\begin{figure}[!htbp]
\includegraphics[width=0.96\columnwidth]{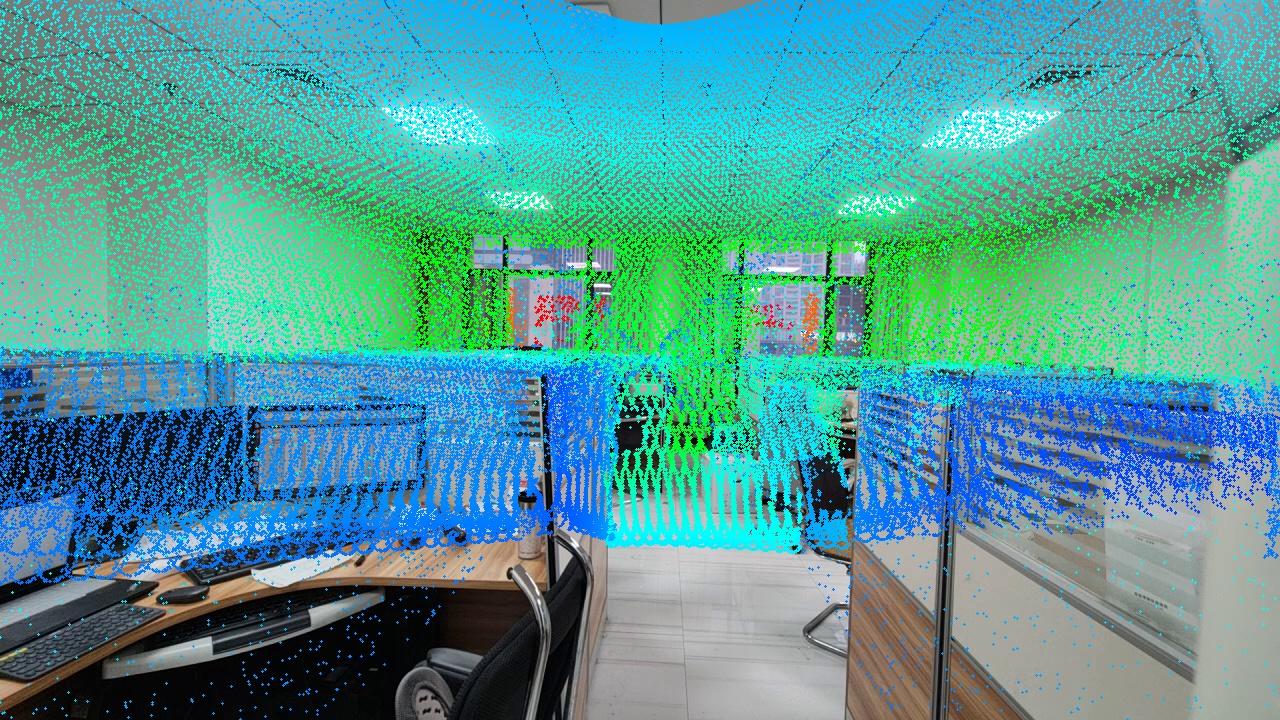}
\caption{
Two seconds of stationary LiDAR scans are projected onto the corresponding S22+ phone image. The LiDAR point colors, derived from depth, exhibit distinct changes that align well with image edges, indicating precise camera-LiDAR calibration.
}
\label{fig:overlay}
\end{figure}

To quantitatively assess the lidar-camera extrinsic calibration, we project an aggregated stationary lidar scan onto the corresponding image. As shown in Fig.~\ref{fig:overlay} for the S22+, the color of the lidar points (which encodes depth) changes distinctly at depth discontinuities, indicating accurate extrinsic calibration between the camera and the lidar.

\subsection{Dataset and Reference}
To evaluate tracking and mapping accuracy, we collected two sequences in a basement parking lot and two outdoor sequences around the Xinghu building using the S22+ device, and a corridor sequence using the K60Pro device. For the outdoor sequences, GNSS-RTK positions with unix times, solved with RTCM data, were recorded from the GNSS stick.

To create a ground truth map, we collected several scans with the Leica RTC360 terrestrial laser scanner (TLS) at both test sites: the parking lot and the road surrounding the building. The TLS scans were registered using the Cyclone software, and the relative poses were further refined with classic point-to-plane ICP \citep{zhouOpen3D2018} and checked by visual inspection.

To generate the reference odometry solution (\ie, the lidar poses in the TLS frame), we employed the Fast-LIO2 \citep{xuFASTLIO2FastDirect2022} odometry method in localization mode, using the TLS map for localization and disabling its incremental mapping. 
To initialize Fast-LIO2 in localization mode, we obtained the starting pose for each sequence by first creating a point cloud map from Fast-LIO2, then aligning this odometry map manually to a specific TLS scan in \texttt{CloudCompare}\footnote{\url{https://www.cloudcompare.org/doc/wiki/index.php/Alignment_and_Registration}}.

\subsection{Tracking Accuracy}
To evaluate the tracking accuracy with the S22+ sequences, we compared several lidar-based odometry methods, including the online Faster-LIO method running on the phone, the lidar-only odometry method KISS-ICP \citep{vizzo2023ral}, the lidar-inertial odometry method Fast-LIO2 \citep{xuFASTLIO2FastDirect2022},
and the lidar-inertial-visual odometry method Fast-LIVO2 \citep{zhengFastLIVO22025},
against reference trajectories for both the basement and building sequences.

The resulting trajectories were first aligned to the TLS reference using an SE3 transformation computed with the Umeyama algorithm. We then calculated the relative pose errors \citep{zhangTutorialQuantitativeTrajectory2018} in translation and rotation across the sequences and computed the root mean square errors in translation and rotation for each sequence.

As shown in Tables \ref{tab:rel-errors} and \ref{tab:abs-errors}, the online Faster-LIO method achieves remarkable accuracy, nearly matching the offline performance of Fast-LIO2. 
The lidar-only method, KISS-ICP, shows slightly lower accuracy compared to the lidar-inertial odometry methods.
Notably, the Fast-LIVO2 method performed worse than the LIO methods across these sequences.
We attribute this to the rolling shutter effect in smartphone cameras, as Fast-LIVO2 assumes a global shutter.

\begin{table}[!ht]
\centering
\caption{
Relative pose errors across four sequences captured by the S22+ device for four odometry methods, without loop closure.}
\label{tab:rel-errors}
\resizebox{0.8\columnwidth}{!}{ 
\begin{tabular}{@{}lcc@{}}
\toprule
           & Translation (\%) & Rotation ($^\circ$/m) \\ \midrule
Faster-LIO & 0.354 &  0.007                  \\ \midrule
Fast-LIO2   & 0.261 &  0.006                 \\ \midrule
KISS-ICP   & 1.029 &  0.026                  \\ \midrule
Fast-LIVO2  &  0.807 &  0.023                      \\ \bottomrule
\end{tabular}
}
\end{table}

\begin{table}[!htbp]
\centering
\caption{Root mean square errors in translation (cm) and rotation ($^\circ$) across four S22+ sequences.}
\label{tab:abs-errors}
\resizebox{1.0\columnwidth}{!}{ 
\begin{tabular}{l|c|c|c|c}
\toprule
& basement1 & basement2 & building1 & building2 \\
\midrule
Faster & 8.1, 0.63 & 10.6, 0.69 & 17.1, 0.41 & 23.4, 0.44 \\
LIO2 & 6.3, 0.45 & 9.5, 0.59 & 22.0, 0.39 & 29.9, 0.45 \\
KISS & 14.8, 1.65 & 14.4, 1.75 & 18.2, 1.18 & 39.2, 1.33 \\
LIVO2 & 10.0, 1.60 & 12.1, 1.47 & 22.6, 1.09 & 36.8, 1.27 \\
\bottomrule
\end{tabular}
}
\end{table}

\begin{figure}[!htbp]
\centering
\includegraphics[width=0.85\columnwidth]{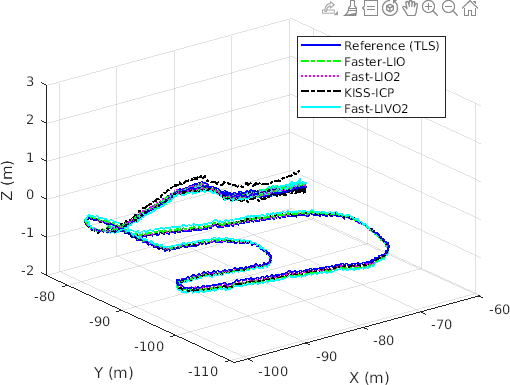}
\includegraphics[width=0.85\columnwidth]{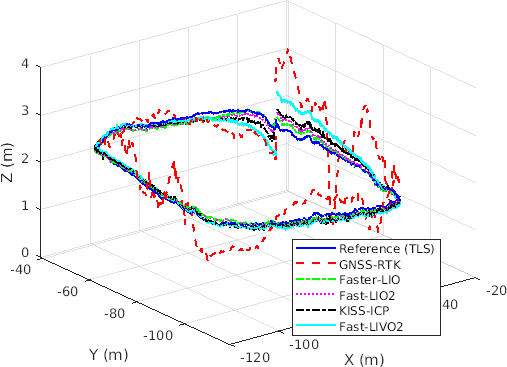}
\caption{The resulting trajectories for the basement1 sequence and the building2 sequence.
Axes scales are not equalized for clearer visualization.
}
\label{fig:trajs}
\end{figure}

The trajectories including the GNSS-RTK trajectory, lidar odometry results, and the reference trajectories are plotted in Fig.~\ref{fig:trajs} for basement-1 and building-2 sequences.
We were pleasantly surprised to find that Faster-LIO exhibited minimal loop closure error, even while running on a resource-constrained smartphone. 
Overall, the data collected with our device can lead to high odometry accuracy.

\subsection{Mapping Accuracy}
Using the above LIO and LIVO approaches, we can get a point cloud map of the test sites. We assess the mapping accuracy of each odometry method by comparing the generated point cloud map to the TLS map.

For each odometry method, we first align the odometry trajectory to the TLS reference trajectory with an SE3 transformation. Next, we undistort and save all scans using the Fast-LIO2 method. With these undistorted scans and the transformed trajectory, we aggregate the scans by transforming each scan with linearly interpolated poses. Finally, we apply random downsampling to reduce the point cloud map size to approximately 20 MB. 
The TLS map is generated by aggregating registered TLS scans and performing random downsampling.

We compare an odometry map with the TLS map in two steps within the CloudCompare software. First, we apply a fine ICP registration 
to better align the two maps.
Second, we compute the cloud-to-cloud distance using local least-squares plane modeling (radius of 0.3 m) with a maximum distance threshold of 0.7 m. The mean distances for the basement and the building sequences are presented in Table~\ref{tab:mean-dist}.

\begin{table}[!htbp]
\centering
\caption{
Mean distances between the point cloud map aggregated from odometry poses and the TLS reference map, calculated using the cloud-to-cloud distance tool in CloudCompare.}
\label{tab:mean-dist}
\begin{tabular}{@{}ccccc@{}}
\toprule
Dist. (cm)  & Faster & LIO2 & KISS & LIVO2 \\ \midrule
basement1 & 27.4 & 27.2 & 28.7 & 32.1 \\
basement2 & 25.0 & 24.8 & 25.3 & 25.6 \\
building1 & 15.7 & 15.7 & 16.5 & 16.5 \\
building2 & 16.7 & 17.9 & 22.3 & 21.2 \\
\bottomrule
\end{tabular}
\end{table}

\begin{figure}[!htbp]
\includegraphics[width=0.37\columnwidth]{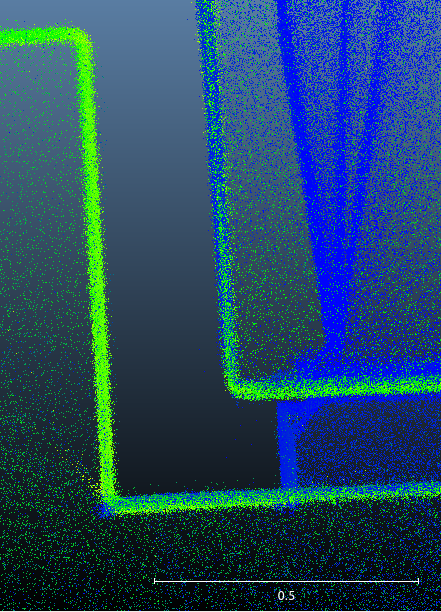}
\includegraphics[width=0.62\columnwidth]{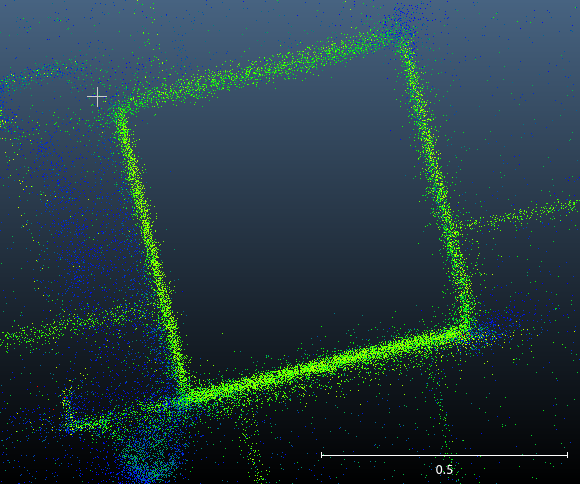}
\caption{
Qualitative evaluation of wall thickness in point cloud maps generated by Fast-LIO2 for the K60Pro corridor sequence (top) and the S22+ basement-1 sequence (bottom). The scale legend in the bottom right of each image indicates that the wall thickness is generally less than 5 cm.}
\label{fig:wall}
\end{figure}

The mean distances between point clouds are often as large as 20 cm, indicating that further mapping improvements are needed.
Through visual inspection of the basement and building mapping results, we observed that wall thickness is generally no more than 5 cm (Fig.~\ref{fig:wall}). However, noticeable point cloud mismatches occur at revisited locations, particularly around features like lampposts.
To address these discrepancies, an immediate refinement would be to apply an offline mapping optimization tool with loop closure, such as BALM2 \citep{liuEfficientConsistentBundle2023}.
Another observed effect is the increased wall thickness in far and/or tall buildings. We attribute this to the decreasing distance accuracy of the Mid360 as range increases. This suggests that the Mid360 may not be well-suited for mapping large areas containing numerous objects at distances exceeding 40 meters.
To address this limitation, integrating a long-range sensor such as the Ouster OS2 with the phone is viable.

\begin{figure}[!htbp]
\includegraphics[width=\columnwidth]{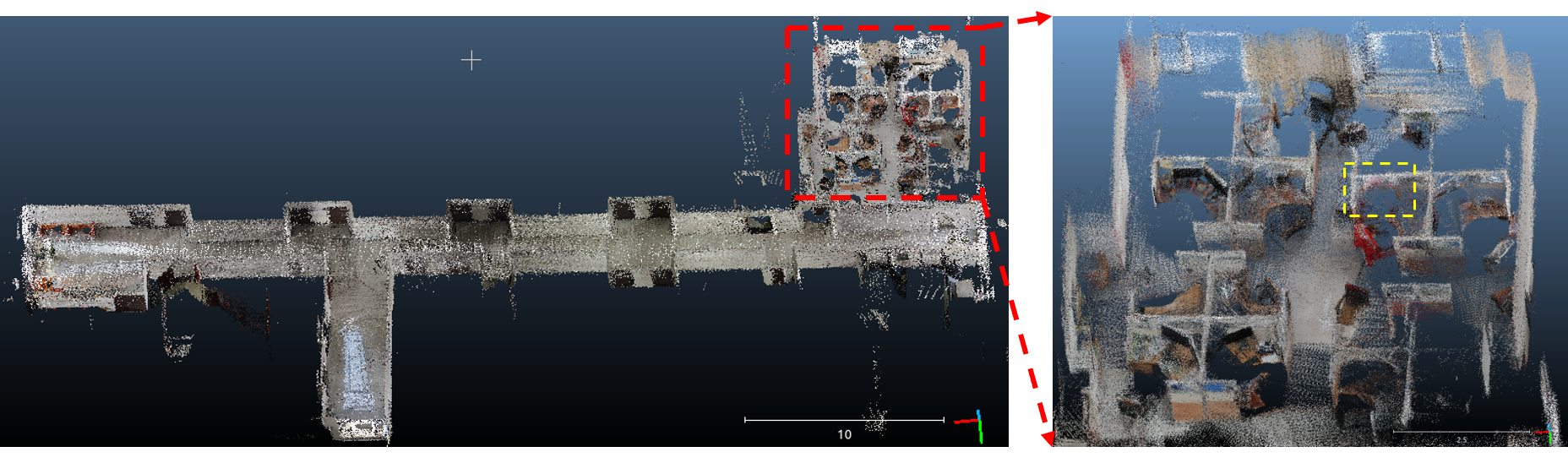} \\
\includegraphics[width=\columnwidth]{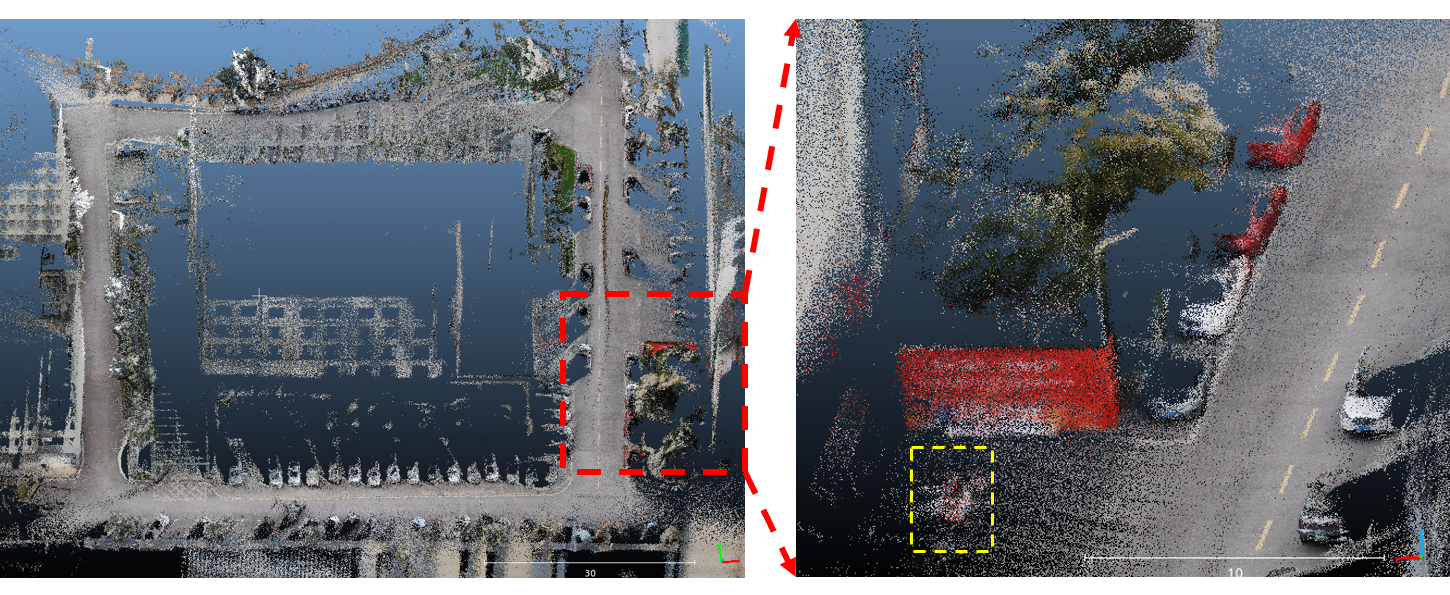}
\caption{
Colored point clouds generated by Fast-LIVO2 for the corridor sequence (top) and the building-2 sequence (bottom).
Note that the zoomed-in view of the building map has a different perspective from the full view.
Also, the building map exhibits noticeable distortion in the top right corner, where a revisit occurs, due to odometry drift.
Minimal color leakage can be seen within the dashed yellow boxes.
}
\label{fig:rgbpcd}
\end{figure}

For qualitative evaluation, Fig.~\ref{fig:rgbpcd} presents the colored point clouds generated by Fast-LIVO2 for the K60 Pro corridor sequence and the S22+ building-2 sequence.
In both cases, Fast-LIVO2 produced visually consistent maps, except at the revisited location. While minor, some color leakage is observed on certain points. Properly addressing the rolling shutter effect should help mitigate this.

\subsection{Runtime Analysis}
We evaluated the runtime performance of Faster-LIO on three smartphone platforms: the K40 Pro (released in 2021), the K60 Pro, and the S22+. The results, presented in Table~\ref{tab:timing}, show that Faster-LIO achieves real-time performance on these devices.
In contrast, Fast-LIO was tested but proved to be at least twice as slow, processing frames at a rate of less than 10 Hz. Additionally, the logging module operates with minimal computational overhead. For example, on the K60 Pro, saving a LiDAR scan takes only 0.43 ms.

\begin{table}[!htbp]
\centering
\caption{Runtime costs of major components of Faster-LIO on smartphones. The frame processing entry is the average processing time for a frame in LIO.}
\label{tab:timing}
\begin{tabular}{@{}cccc@{}}
\toprule
\begin{tabular}[c]{@{}c@{}}Times\\ (ms)\end{tabular} &
  \begin{tabular}[c]{@{}c@{}}Point cloud\\ downsample\end{tabular} &
  \begin{tabular}[c]{@{}c@{}}IEKF\\ update\end{tabular} &
  \begin{tabular}[c]{@{}c@{}}Frame\\ proc.\end{tabular}
  \\ \midrule
K40Pro &
  32.80 &
  14.25 &
  \textbf{66.54} \\ \midrule
K60Pro &
  13.67 &
  24.91 &
  \textbf{64.46} \\ \midrule
S22+ &
  19.79 &
  9.20 &
  \textbf{40.56} \\ \bottomrule
\end{tabular}
\end{table}

\subsection{Limitations}
Our device requires knowledge of the Livox lidar’s IP address each time it reboots. The phone’s Ethernet IP address changes upon reboot, necessitating the use of
the Livox Viewer to reset the lidar’s IP address to match that of the smartphone.
Moreover, battery consumption is noticeably higher when the phone is connected to the GNSS-RTK stick via Bluetooth.


\section{Conclusions}\label{sec:conclusions}
To address the demand for low-cost, ubiquitous mapping, we developed an Android-based mobile mapping system integrating a Livox lidar and a GNSS-RTK stick. The system is built using ROSJava for message dispatching and the NDK for compiling C++-based ROS packages. It supports a real-time LIO module, a data logging module, and an interactive user interface. With the proposed procedures, we demonstrate consistent calibration and repeatable time offsets. By processing the collected lidar-inertial-visual data using recent lidar-based odometry and mapping methods, we showcase the potential of this device for medium-scale mapping applications.

%

{
	\begin{spacing}{1.17}
		\normalsize
		\bibliography{zotero} 
	\end{spacing}
}

\end{document}